\documentclass[ejs]{imsart}

\RequirePackage{amsthm,amsmath,amsfonts,amssymb}
\RequirePackage[numbers]{natbib}
\RequirePackage[colorlinks,citecolor=blue,urlcolor=blue]{hyperref}
\RequirePackage{graphicx}

\usepackage{algorithm}
\usepackage{algorithmic}
\usepackage{bbm}

\startlocaldefs
\theoremstyle{plain}

\newtheorem{theorem}{Theorem}[section]
\newtheorem{lemma}[theorem]{Lemma}
\theoremstyle{definition}

\newtheorem{remark}{Remark}
\theoremstyle{remark}

\newcommand{\norm}[1]{\left\lVert#1\right\rVert}
\endlocaldefs

\begin{document}
\begin{frontmatter}
\title{Efficient learning of differential network in multi-source non-paranormal graphical models}
\runtitle{Efficient learning of differential network}

\begin{aug}
\author[A]{\fnms{Mojtaba}~\snm{Nikahd}\ead[label=e1]{nikahd@ce.sharif.edu}}
\and
\author[A]{\fnms{Seyed Abolfazl}~\snm{Motahari}\ead[label=e2]{motahari@sharif.edu}}
\address[A]{Computer Engineering Department,
Sharif University of Technology\printead[presep={,\ }]{e1,e2}}

\runauthor{M. Nikahd et al.}
\end{aug}

\begin{abstract}
This paper addresses learning of sparse structural changes or differential network between two classes of non-paranormal graphical models. We assume a multi-source and heterogeneous dataset is available for each class, where the covariance matrices are identical for all  non-paranormal graphical models. The differential network, which are encoded by the difference precision matrix, can then be decoded by optimizing a lasso penalized D-trace loss function.
To this aim, an efficient approach is proposed  that outputs the exact solution path, outperforming the previous methods that only sample from the solution path in pre-selected regularization parameters. Notably, our proposed method has low computational complexity, especially when the differential network are sparse. Our simulations on synthetic data demonstrate a superior performance for our strategy in terms of speed and accuracy compared to an existing method. Moreover, our strategy in combining datasets from multiple sources is shown to be very effective in inferring differential network in real-world problems. This is backed by our experimental results on drug resistance in tumor cancers. In the latter case, our strategy outputs important genes for drug resistance which are already confirmed by various independent studies.
\end{abstract}



\end{frontmatter}


\section{Introduction}
Underlying structures that govern the interactions between the components of a large system can sometimes be modeled by graphical models. Graphical models are used extensively in many scientific fields including computational biology \citep{ni2018bayesian}, social sciences \citep{amati2018social}, financial sciences \citep{giudici2018financial}, and  cognitive sciences \citep{zhang2018aberrant} among others. In many applications, it is pivotal to recover any structural changes between two different states of a system. In computational biology for instance, changes of regulatory mechanisms of normal cells and cancer cells reveal disease mechanisms \citep{west2012differential}. Therefore, efficient methods for detecting and analyzing structural changes in graphical models have garnered significant attention.

Various methods have been developed to address the identification of structural changes, commonly referred to as differential network analysis. A comprehensive review of these methods can be found in \citep{shojaie2020differential}. Gaussian Graphical Models (GGMs) are prominent in modeling structures \citep{mazza2020learning, tavassolipour2018learning}, particularly in the modeling of  biological networks \citep{ou2018joint,yuan2017differential}. Under this model, the inference of structures can be obtained by determining the sparsity pattern of the precision matrix \citep{friedman2008sparse}.

Several approaches have been proposed to detect differential network of two GGMs. In the first approach, the precision matrix of each group is estimated separately and immediately used to infer changes in the underlying structures. Any method that is proposed for estimating the precision matrix of a GGM  can be used with this strategy \citep{meinshausen2006high,friedman2008sparse,cai2011constrained,zhang2014sparse}. Methods of the second approach jointly estimate the precision matrices and their differences \citep{guo2011joint,mohan2014node}. One caveat of both the first and second approaches is that the strong assumption of the sparsity of each  precision matrix is imposed on the model. However, if only the differential network are to be recovered then there is no need to recover individual precision matrices yet putting the strong assumption on the matrices. 
In the third approach, the differential network are directly estimated without any assumption on the sparsity of the precision matrices \citep{zhao2014direct,tian2016identifying,yuan2017differential,zhang2017incorporating,he2019direct,tang2020fast,na2021estimating}. This approach offers more flexibility and allows for the identification of changes in the graphical structure without explicitly estimating the complete precision matrices. By relaxing the sparsity assumption, these methods provide a means to focus solely on capturing the differences between the GGMs.

Taking the third approach, \cite{zhao2014direct} introduced a new  estimator which is shown to be consistent under mild conditions by assuming that the differential network are sparse. However, the proposed method \citep{zhao2014direct} is time-consuming and requires a surplus step to symmetrize the estimated differential matrix. In another work, the minimization of the lasso penalized D-trace loss function that directly infers the differential network is proposed \citep{yuan2017differential}. \cite{yuan2017differential} exploited the D-trace loss function \citep{zhang2014sparse} and introduced a one-step symmetric estimator. It is shown that the estimator proposed by \cite{yuan2017differential} is consistent under milder conditions than the conditions introduced by \cite{zhao2014direct}.


To solve the minimization of the lasso penalized D-trace loss function, an iterative algorithm based on the alternating direction method of multipliers (ADMM) \citep{boyd2011distributed} is developed by \cite{yuan2017differential}. The computational complexity of the algorithm is $\mathcal{O}\left( d^3 \right)$ (where $d$ is the data dimension) in each iteration. Recently, a new fused D-trace loss function is proposed to estimate the differential network by \cite{wu2020differential}.
Similar to \cite{yuan2017differential}, an iterative method is introduced to solve the new optimization problem with $\mathcal{O}\left( d^3 \right)$ computations for each iteration. Numerically, it is shown that the method proposed by \cite{wu2020differential} outperforms that of \cite{yuan2017differential}. So when $d$ is relatively large especially in the high-dimensional setting, the method will be computationally prohibitive. To tackle this challenge,  \cite{tang2020fast} considered the model defined by \cite{yuan2017differential} under a special setting and proposed a new iterative method to approximate the solution. The computational complexity of each iteration of the proposed algorithm is $\mathcal{O} \left(nd^2\right)$, where $n$ is the sum of the samples in both categories.

Despite significant research advancements, there are still challenges associated with the practical application of these methods to real-world problems. One prominent issue is the computational cost, as previously mentioned. In addition to this concern, the ability to effectively utilize all available datasets for each subject is another critical consideration in the development of these methods. The requirement for this issue comes from the fact that it may be prohibitively expensive for a single expert to create a high-quality dataset and at the same time, there are different datasets produced by different experts. Hence, it is very convincing to integrate the data from various sources. This is a common practice, particularly in biological applications \citep{zhang2016differential,ou2017node,ou2018joint,ou2021wdne}. It is important to note that datasets obtained from various sources vary greatly. Especially, genetic data gathered from multiple laboratories may have been obtained via different measurement techniques \citep{wahlsten2003different}. Therefore, samples obtained from diverse sources  may not be modeled by a GGM. To make the model more flexible, non-paranormal graphical models have been introduced \citep{liu2009nonparanormal}. These models provide enough freedom to consider a unique underlying structure for datasets yet they allow considering the heterogeneity of datasets due to sampling conditions \citep{liu2012high,liu2009nonparanormal,xu2016semiparametric,zhang2017incorporating}.

In this paper, we propose an efficient method for extracting differential network between two non-paranormal graphical models in high-dimensional settings where the changes are assumed to be $s$-sparse.  No assumptions are imposed on the sparsity of individual structures. To overcome the computational challenges, we propose a new iterative algorithm that minimizes the lasso penalized D-trace loss function for a specific interval of regularization parameter values in a cost-effective manner. Besides being efficient in terms of speed and accuracy, our proposed method is applied to real datasets where differential network are  successfully discovered. We will elaborate on each of these contributions next.

In terms of speed and in the high-dimensional setting where the number of structural changes is sparse, our proposed method is faster than the algorithm proposed by \cite{yuan2017differential}. This is because the algorithm \citep{yuan2017differential} performs many computations for identifying the candidate structural change-sets in the parameter tuning step. More concretely, the algorithm \citep{yuan2017differential} uses $\mathcal{O} \left(kd^3\right)$ computations (where $k$ is the number of iterations and controls the approximation error of the algorithm) to generate a candidate while our proposed method identifies a unique candidate at each iteration with only $\mathcal{O} \left(cd^2\right)$ computations, where $c$ is a parameter that controls the maximum number of non-zero elements in the solution path and $c \ll d$. In addition, our method is faster than the method proposed by \cite{wu2020differential}. This is due to the fact that \cite{wu2020differential} and \cite{yuan2017differential} have similar computational complexity.
Likewise, in the regime $n>d$,
our method achieves lower computational complexity than the algorithm proposed by \cite{tang2020fast}. We compare our proposed method with the methods proposed in \cite{yuan2017differential} and \cite{wu2020differential} on synthetic data to illustrate the time efficiency of our approach.

In terms of accuracy, our proposed approach, unlike previous methods that approximate the solution, identifies the exact solution thus eliminating the approximation error. In addition, datasets are assumed to be heterogeneous, which means they are provided from different sources. We propose a data integration method and show it has the same order of sample complexity as the one shown in \cite{yuan2017differential} and \cite{wu2020differential}
under the assumption that datasets are homogeneous. In other words, our proposed method obtains flexibility without sacrificing the sample complexity. Also, This result is justified by a thorough analysis and several numerical experiments on synthetic datasets.

In terms of applications, our proposed method is evaluated in a real biological application where the differential network between drug-resistant and drug-sensitive cases in cancer patients are under investigation. Several gene expression datasets are incorporated in both groups and our method is applied. Our proposed method properly identifies the genes involved in the drug response mechanism and the result is evaluated by a relevant validated gene list.



The rest of this paper is organized as follows. In Section \ref{sec:Materials and Methods}, we present the model, our optimization algorithm, convergence, and consistency analysis. In Sections \ref{sec:Real data application} and \ref{sec:Real data application}, we provide our numerical experiment results for the simulated and real datasets.  Section \ref{sec:Conclusion} concludes the paper.


\subsection{Notations}
The set $\{1,\cdots,m\}$ is represented by $[m]$.
For a matrix $A \in \mathbb{R}^{d \times d}$, we denote the sub-matrix of $A$ whose rows are in a set $\mathcal{R} \subseteq \left[d\right]$ and columns are in a set $\mathcal{C} \subseteq \left[d\right]$ as $A_{\mathcal{R},\mathcal{C}}$.
We refer to $i$th row and the $j$th column of $A$ by $A_{i,:}$ and $A_{:,j}$, respectively. Similarly, for a vector $\mathbf{x}$ and a set $\mathcal{S}$, we write $\mathbf{x}_{\mathcal{S}}$ to denote a sub-vector of $\mathbf{x}$ whose components are in $\mathcal{S}$.
Let
$\norm{A}_{1} = \sum_{i,j=1}^{d} \left| A_{i,j} \right|$ and $\norm{A}_{\infty} = \max_{i,j}
\left| A_{i,j} \right|$ denote the element-wise $\ell_1$ and max norm, respectively. $\textrm{vec}\left(A\right)$ denotes the vectorized form of matrix $A$ which is a $d^2 \times 1$ vector obtained by stacking the columns of $A$. In addition, the inner product of two matrices is defined as $\langle A, B \rangle = \textrm{tr}\left(AB^{\intercal}\right)$. 
For a set $\mathcal{S}$, the cardinality of $\mathcal{S}$ is denoted as $|\mathcal{S}|$.

\section{Material and methods}
\label{sec:Materials and Methods}

In this section, we propose a method for learning the differential network of two distinct, yet structurally similar, non-paranormal graphical models. Data samples generated from each of the two models are also heterogeneous, meaning that except for the underlying structure, the distributions of the samples are allowed to be different. Our method directly discovers the differential network without estimating the structure of the two models. To this end, we first  give a brief review on non-paranormal models.

\subsection{Non-paranormal graphical models}
The non-paranormal graphical model (or the Gaussian copula distribution) is an extension of the GGM described as follows. A d-dimensional random vector $\mathbf{x} = {\left[x_{1}, x_{2}, \dots, x_{d}\right]}^{\intercal}$ is said to follow a non-paranormal distribution, denoted by $\mathbf{x} \sim \mathrm{NPN}\left(\mathbf{0},\Sigma,\mathbf{f}\right)$, if there exists a set of univariate monotonically-increasing functions $\mathbf{f} \triangleq{\left\lbrace f_{j}\right\rbrace}_{j=1}^{d}$ such that the transformed random vector $\left[ f_1\left(x_1\right), \cdots, f_d\left(x_d\right) \right]^{\intercal}$ follows a multivariate Gaussian distribution with zero mean and covariance matrix $\Sigma$. The precision matrix is denoted by  $\Sigma^{-1}$. In \cite{liu2009nonparanormal}, it has been shown that the conditional independence structure of $\mathbf{x}$ is encoded by the sparsity pattern of the precision matrix. In other words,  random variables $x_i$ and $x_j$ are conditionally independent given all other variables if and only if ${\left[\Sigma^{-1}\right]}_{i,j}=0$.
This property can be used to infer  the graphical structure of $\mathbf{x}$ by obtaining an estimate of the sparsity pattern of $\Sigma^{-1}$.

It has been shown that there exists a near optimal parameter estimation and a graph recovery algorithm for non-paranormal models, which has the same sample complexity bound as that of an optimal Gaussian model estimation method \citep{liu2012high,xue2012regularized}. Based on these results, authors have suggested the non-paranormal graphical model can be considered as a safe substitute for the widely used multivariate Gaussian  model, even when the samples are truly normal.

\subsection{Problem setting}

For a given covariance matrix 
$Q$ and a set of univariate monotonically-increasing functions
$\mathbf{f} = \{ f_j \}^d$, a dataset $\mathcal{D} \triangleq \left\lbrace \mathbf{x}^{(1)}, \mathbf{x}^{(2)}, \dots, \mathbf{x}^{(m)} \right\rbrace$ is defined as a set of samples
drawn from a non-paranormal with parameters  
$\left(\mathbf{0},Q,\mathbf{f}\right)$. In our setting, the set of functions $\mathbf{f}$ is unknown. This is a challenging property of some real datasets. In particular, in gene expression data analysis, identifying the appropriate data transformation function is an important challenge addressed by \cite{zwiener2014transforming}.
$m$ is called the sample size of the dataset.
A collection of datasets $\left\lbrace \mathcal{D}_{s} \right\rbrace_{s=1}^{N}$ is called heterogeneous if  all the datasets have similar covariance matrices but they possess different functions and sample sizes.

In our problem, two heterogeneous datasets $\left\lbrace \mathcal{D}_{s} \right\rbrace_{s=1}^{N}$ and
$\left\lbrace \mathcal{D}'_{s'} \right\rbrace_{s'=1}^{N'}$ with covariance matrices $\Sigma,\Sigma' \in \mathbb{R}^{d \times d}$ are given. 
We assume, without loss of generality, that the diagonal elements of each covariance matrix are set to 1. Our goal is to discover the non-zero entries of the difference of precision matrices which encodes the structural changes between the two models. More concretely, let $\Delta^*$ denote the sparse matrix that corresponds to the difference between precision matrices of the two models, i.e.,
$\Delta^* = \left(\Sigma\right)^{-1} - \left(\Sigma'\right)^{-1}\in\mathbb{R}^{d\times d}$. Suppose 
$\mathcal{A}^* = \left \lbrace
\left(i,j\right) : \Delta^*_{i,j} \neq 0
\right \rbrace$ be the support of $\Delta^*$ and $s=\left|\mathcal{A}^*\right|$. Also, we assume that the true differential network $\Delta^*$ is sparse, i.e., $s \ll d$. Our goal is to extract the support of the differential network  $\Delta^*$ through processing the information within all the datasets.

To directly infer $\Delta^*$, we use the lasso penalized D-trace model proposed by \cite{yuan2017differential}. This model has a regularization parameter to control the sparsity of the result. In this paper, we present an efficient algorithm to find the solution path of this optimization problem. In other words, instead of traditional methods that solve the problem for a specific value of the regularization parameter, our algorithm extracts the solution of the problem for any values of the regularization parameter in a specific interval. Also, we show that our estimator is consistent and the sample complexity of our method has the same order as the one shown by \cite{yuan2017differential} despite the flexibility provided in our problem setting.

\subsection{Algorithm}

The lasso penalized D-trace model, introduced by \cite{yuan2017differential}, utilizes estimate of covariance matrices denoted by $\hat{\Sigma}$ and $\hat{\Sigma}'$. The model is formulated as the following optimization problem
\begin{equation}
\label{lasso penalized D-trace loss function minimization}
\min_{\Delta}
L_D\left(\Delta;
\hat{\Sigma},
\hat{\Sigma}'\right) +
\lambda \norm{\Delta}_{1},
\end{equation}
where $\lambda > 0 $ is a tuning parameter, and 
$L_D\left(\Delta;
\hat{\Sigma},
\hat{\Sigma}'\right)$ represents the D-trace loss function defined as
\begin{align}
\label{D-trace loss function}
\frac{1}{4}
\left(
\langle
\hat{\Sigma}\Delta,
\Delta\hat{\Sigma}'
\rangle +
\langle
\hat{\Sigma}'\Delta,
\Delta\hat{\Sigma}
\rangle
\right) -
\langle
\Delta,
\hat{\Sigma} - \hat{\Sigma}'
\rangle.
\end{align}

The Hessian of the function \eqref{D-trace loss function}
with respect to $\Delta$ is 
$\left(
\hat{\Sigma} \otimes \hat{\Sigma}' +
\hat{\Sigma}' \otimes \hat{\Sigma}
\right)/2$, where $\otimes$ denotes the Kronecker product. Therefore, given that both $\hat{\Sigma}$ and $\hat{\Sigma}'$ are positive semi-definite matrices, the optimization problem \eqref{lasso penalized D-trace loss function minimization} is a convex program. Consequently, there is a threshold $\lambda_{T} \ge 0$, 
such that the problem \eqref{lasso penalized D-trace loss function minimization} has a unique solution for any $\lambda \in \left(\lambda_T,\infty\right)$.
Hence, any $\Delta$  satisfying the subgradient optimality conditions is an optimal solution for the problem  \eqref{lasso penalized D-trace loss function minimization}. From the subgradient optimality conditions, it is possible to identify the solution path as a function of tuning parameter $\lambda \in \left(\lambda_T,\infty\right)$, which we will write as $\hat{\Delta}\left(\lambda\right)$. The obtained solution path is a continuous piecewise linear function. The precise definition of this property is stated in the following theorem and the proof is presented in \ref{sec:Proof of Theorem piecewise linear solution path}.

\begin{theorem}
	\label{piecewise linear solution path}
	There are $\infty = \lambda_0 > \lambda_1 > \lambda_2 > \dots > \lambda_T \ge 0$, such that $\hat{\Delta}\left(\lambda\right)$ for any $\lambda \in \left(\lambda_T,\infty\right)$ is given by  
	\begin{equation*}
	\frac{\lambda_{t} - \lambda}{\lambda_{t} - \lambda_{t+1}} \hat{\Delta} \left( \lambda_{t+1} \right) +
	\frac{\lambda - \lambda_{t+1}}{\lambda_{t} - \lambda_{t+1}} \hat{\Delta} \left( \lambda_{t} \right), \forall \lambda \in \left[\lambda_{t+1}, \lambda_t\right].
	\end{equation*}
\end{theorem}

Based on Theorem \ref{piecewise linear solution path}, we propose an efficient algorithm to recover all the knots in the solution path. In fact, knots are points joining two adjacent line segments in the solution path. The knots are denoted by $\lambda_t$ for  $t=0,1,\dots,T$. For notational simplicity,  The proposed method is given in Algorithm \ref{alg:solution path algorithm}. The algorithm is obtained by exploiting the sparsity of the solution in the subgradient optimality conditions. Our proposed algorithm has two termination conditions.
One condition ensures uniqueness, while the other controls the sparsity of the solutions.
The uniqueness condition is verified by satisfying the uniqueness of the necessary conditions derived for Problem \eqref{lasso penalized D-trace loss function minimization}. 
Moreover, checking the sparsity level helps in reducing computation time as it inhibits the algorithm from falling into a dense and improper structure.
To see the detail of the computations, the reader is referred to the proof of Theorem \ref{piecewise linear solution path} which is presented in \ref{sec:Proof of Theorem piecewise linear solution path}.

\begin{algorithm}[t!]
\caption{Lasso Penalized D-trace Loss Function Minimization Path}\label{alg:solution path algorithm}
\algsetup{linenosize=\tiny}
  \scriptsize
\begin{algorithmic}[1]

\REQUIRE maximum number of non-zero elements in the solution path $c>0$ and estimated covariance matrices $\hat{\Sigma} \succeq 0$ and $\hat{\Sigma}' \succeq 0$

\STATE Initialize iteration counter $t=0$, tuning parameter $\lambda_0=\infty$, active set $\mathcal{A}=\emptyset$, and active signs $\mathbf{s}_{\mathcal{A}}=\emptyset$.

\WHILE{$\left|\mathcal{A}\right| \leq c \And \Gamma_{\mathcal{A}, \mathcal{A}} \text{ is invertible}$}
    \STATE Compute $\Gamma_{:,\mathcal{A}}$, 		$\left(\Gamma_{\mathcal{A},\mathcal{A}}\right)^{-1}\mathbf{v}_{\mathcal{A}}$ and	$\left(\Gamma_{\mathcal{A},\mathcal{A}}\right)^{-1}\mathbf{s}_{\mathcal{A}}$.

    \STATE Compute the next hitting time, where $\overset{+}{\max}$ denotes the maximum of those arguments that are less than $\lambda_t$.
		\begin{equation*}
		    \lambda_{t+1}^{hit}=\overset{+}{\max_{e\in \mathcal{A}^{c},s_e\in\left\lbrace-1,1\right\rbrace}}
		\frac{
			\Gamma_{e,\mathcal{A}}
			\left(\Gamma_{\mathcal{A},\mathcal{A}}\right)^{-1}
			\mathbf{v}_{\mathcal{A}} - 
			V_{e}
		}{s_e - 
			\Gamma_{e,\mathcal{A}}
			\left(\Gamma_{\mathcal{A},\mathcal{A}}\right)^{-1}
			\mathbf{s}_{\mathcal{A}}
		}.
		\end{equation*}
		
	\STATE Compute the next crossing time,
		\begin{equation*}
		    \lambda_{t+1}^{cross}=\overset{+}{\max_{e\in \mathcal{A}}} \frac{
			-
			\left[
			\left(\Gamma_{\mathcal{A},\mathcal{A}}\right)^{-1}
			\mathbf{v}_{\mathcal{A}}
			\right]_{e}
		}{
			\left[
			\left(\Gamma_{\mathcal{A},\mathcal{A}}\right)^{-1}
			\mathbf{s}_{\mathcal{A}}
			\right]_{e}
		}.
		\end{equation*}

	\STATE Compute the next knot time
	\begin{equation*}
	    \lambda_{t+1} = \max \left\lbrace
		\lambda_{t+1}^{hit}, \lambda_{t+1}^{cross}
		\right\rbrace.
	\end{equation*}
		
	\STATE Compute the solution as $\lambda$ decreases from $\lambda_t$ until $\lambda_{t+1}$ by
	\begin{align*}
	    &\textrm{vec}\left(\hat{\Delta}\left(\lambda\right)\right)_{\mathcal{A}} = -
	\left(\Gamma_{\mathcal{A},\mathcal{A}}\right)^{-1}
	\left(
	\mathbf{v}_{\mathcal{A}} +
	\lambda \mathbf{s}_{\mathcal{A}}
	\right) \\
	    &\textrm{vec}\left(\hat{\Delta}\left(\lambda\right)\right)_{\mathcal{A}^{c}} = 0.
	\end{align*}
	
	\IF{$\lambda_{t+1}^{hit} > \lambda_{t+1}^{cross}$}
	    \STATE Add the hitting variable to $\mathcal{A}$ and its sign to $\mathbf{s}_{\mathcal{A}}$.
	\ELSE
	    \STATE Remove the crossing variable from $\mathcal{A}$ and its sign from $\mathbf{s}_{\mathcal{A}}$.
	\ENDIF
	
	\STATE Update $t=t+1$.
\ENDWHILE

\ENSURE Solution path to differential network estimator $\hat{\Delta}\left(\lambda\right)$
\end{algorithmic}
\label{alg1}
\end{algorithm}

The computational complexity of the proposed algorithm is  $\mathcal{O}(cd^2+c^3)$ per iteration where the number of iterations
is controlled by the maximum number of non-zero elements in the solution path denoted by $c$. To see this, we write
$\Gamma = \frac{1}{2} \left\lbrace \hat{\Sigma} \otimes \hat{\Sigma}' +
\hat{\Sigma}' \otimes \hat{\Sigma} \right\rbrace$
,
$\mathbf{v} = \textrm{vec}\left(
\hat{\Sigma} - \hat{\Sigma}'
\right)$,
and
$\mathcal{A}^{c} = \left\lbrace 1,\dots, d^2\right\rbrace \setminus \mathcal{A}$. In each iteration, $\Gamma_{:,\mathcal{A}}$ and
$\left(\Gamma_{\mathcal{A},\mathcal{A}}\right)^{-1}$ are computed by $\mathcal{O}(cd^2)$ and $\mathcal{O}(c^3)$ computations, respectively. Also, the calculation of Step 4 in Algorithm 1 requires $\mathcal{O}(cd^2)$. The other steps in each iteration require $\mathcal{O}(c)$ computations. Thus, the computational complexity of each iteration is $\mathcal{O}(cd^2+c^3)$.

\subsection{Covariance matrix estimation}
\label{sec:covariance matrix estimation}
We focus on the covariance estimation of the first collection of datasets. The other collection can be treated similarly. Recall that the diagonal elements of  each covariance matrix have been assumed to be $1$. Therefore, each covariance matrix is equal to its corresponding Pearson correlation matrix. In addition, we assumed the set of transformation functions $\mathbf{f}$ is unknown. Therefore, we cannot directly estimate the covariance matrix of the underlying GGM. However, for multivariate Gaussian models, it has been already shown that there exists a one-to-one mapping between the Pearson correlation  and Kendall's tau correlation
\citep{kruskal1958ordinal,kendall1948rank}.
The same relationship has been also established for the case of non-paranormal distributions.
In other words, if
$ \mathbf{z} \sim \mathrm{NPN}\left(\mathbf{0}, \Sigma, \mathbf{f}\right)$ represents a multivariate non-paranormal model of dimension $d$,
then for $k,l\in\left[d\right]$, we have
\begin{equation}
\label{eq:kendalltauproperty}
\Sigma_{k,l} = \sin \left(\frac{\pi}{2}\tau_{k,l}\right),
\end{equation}
where $\tau_{k,l}$ denotes Kendall's tau correlation between $z_k$ and $z_l$. We exploit this relationship to develop our Pearson correlation estimator.

For a given dataset $\mathcal{D}_s$, data points can be represented by a matrix $X$ of size $m_s\times d$.
The empirical estimate of Kendall's tau between the $k$th and $l$th dimensions is
\begin{align}
\label{eq:kendalltaudef}
&\hat{\tau}^{(s)}_{k,l} =2
\frac{\sum_{1\le i < j \le m_{s}}
\mathrm{sign}\left[
\left( X_{i,k}-X_{j,k} \right)
\left( X_{i,l}-X_{j,l} \right)
\right]}{m_s\left(m_s-1\right)} 
.
\end{align}

To obtain an estimate of the covariance matrix for a collection of datasets, we use the following statistics
\begin{equation}
\hat{\tau}^{(w)}_{k,l} = \sum_{s=1}^{N} \frac{m_s}{m} \hat{\tau}^{(s)}_{k,l},
\end{equation}
where $m=\sum_{s=1}^{N} m_s$ and $N$ is the number of datasets. The ratio $\frac{m_s}{m}$ is chosen to minimize a theoretical upper bound on the absolute value of estimation error which is detailed in Lemma \ref{errorBoundOfCorrelationEstimator} in \ref{lem:app_proof}. Using \eqref{eq:kendalltauproperty}, an empirical version of correlation matrix $\tilde{\Sigma}_{k,l}$ can be considered as
\begin{equation}
\tilde{\Sigma}_{k,l} =
\begin{cases}
\sin\left(\frac{\pi}{2}\hat{\tau}^{(w)}_{k,l}\right), & k\neq l\\
1, & k=l.
\end{cases}
\label{Greiner's equality}
\end{equation}

The KKT conditions are sufficient for optimality of the problem \eqref{lasso penalized D-trace loss function minimization} if the covariance matrix estimators are positive semi-definite. Unfortunately,
Kendall's tau estimator doesn't guarantee positive semi-definiteness. Therefore, we need to project  $\tilde{\Sigma}$ onto the cone of positive semi-definite matrices. For this purpose, we use the method proposed by \cite{zhao2014positive}. The positive semi-definite surrogate of the estimated covariance matrix is denoted by $\hat{\Sigma}$. It can be seen that the proposed statistics $\hat{\Sigma}$ is a consistent estimator of the covariance matrices $\Sigma$.

We would like to obtain an upper bound on the sample complexity of our proposed algorithm. To this end, we follow the methodology used by \cite{yuan2017differential} that results in  a sample complexity bound for the sub-Gaussian distribution. In fact, it suffices to modify a few steps of the proof presented in \cite{yuan2017differential} for Theorems 1 and 2 to derive our result.

To this end, we only need to provide the satisfaction of the tail conditions for our proposed covariance matrices estimators $\hat{\Sigma}$ and $\hat{\Sigma}'$. In Lemma \ref{errorBoundOfCorrelationEstimator}, the satisfaction of tail conditions is proved for our proposed covariance matrices estimators $\hat{\Sigma}$ and $\hat{\Sigma}'$.

\begin{lemma} 
	\label{errorBoundOfCorrelationEstimator}
	For any $t>0$ and for all $\left(k,l\right) \in [d] \times [d]$, we have
	\begin{align*}
	&\mathbb{P}\left(
	\left|\tilde{\Sigma}_{k,l} - {\Sigma}_{k,l} \right| \ge t
	\right) 
	\le
	\exp \left(
	\frac{- m t^2}{2 \pi^2}
	\right), \\
	&\mathbb{P}\left(
	\left|{\tilde{\Sigma}'}_{k,l} - {{\Sigma}'}_{k,l} \right| \ge t
	\right) 
	\le
	\exp \left(
	\frac{- m' t^2}{2 \pi^2}
	\right).
	\end{align*}
\end{lemma}
See \ref{lem:app_proof} for the proof.

To state our theorems, we define

\begin{align*}
&\Gamma^* = \frac{1}{2} \left\lbrace \Sigma \otimes \Sigma' +
\Sigma' \otimes \Sigma \right\rbrace,
\alpha = 1- \max_{e \in \mathcal{A}^{*^{c}}} \norm{\Gamma^*_{e,\mathbf{A^*}}
    \left(
    	\Gamma^*_{\mathcal{A}^*, \mathcal{A}^*}
    \right)^{-1}
    }_{1},
G_B=\sqrt{mm'}, \\
&G_A = m^{-1/2} + {m'}^{-1/2},
    \kappa_{\Gamma} =\norm{\left(
    	\Gamma^*_{\mathcal{A}^*, \mathcal{A}^*}
    \right)^{-1}}_{1, \infty},
    \tilde{M} = 24s \kappa_{\Gamma}\left( 2s \kappa_{\Gamma} + 1\right)/\alpha, \\
&M\left( \Delta \right) = \left\lbrace \textrm{sign}\left( \Delta_{j,k} \right) : j,k=1,\dots,d \right \rbrace,
\gamma = \pi\sqrt{2 \left(\eta \log{d} + \log{2} \right)}. 
\end{align*}

\begin{theorem} \label{errorBoundofDeltaEstimator}
If 
$\max_{e \in \mathcal{A}^{*^{c}}} \norm{\Gamma^*_{e,\mathbf{A^*}}
    \left(
    	\Gamma^*_{\mathcal{A}^*, \mathcal{A}^*}
    \right)^{-1}
    }_{1} < 1$,
    $\min \left[ m, m' \right] > \frac{25}{4}  \bar{\bar{\delta}}^{-2} \gamma^2$ and 
    $\lambda_n = 
    \gamma \max \left[
    \frac{2\left( 4-\alpha \right) G_A}{\alpha},
    \tilde{M}
    \left(
    \gamma G_B + G_A
    \right)
    \right]$
    for some $\eta > 2$ then, with probability greater than $1- 2/d^{\eta - 2}$, the support of $\hat{\Delta}$ lies in the support of $\Delta^*$ and
    \begin{align*}
    &\norm{\hat{\Delta} - \Delta^*}_{\infty} \leq M_G \sqrt{\frac{\eta \log d + \log 2}{\min \left[m, m'\right]}}, \\
    &\norm{\hat{\Delta} - \Delta^*}_{F} \leq  M_G \sqrt{s\frac{\eta \log d + \log 2}{\min \left[m, m'\right]}},
    \end{align*}
where $M_G$ and $\bar{\bar{\delta}}$ are constants depending on $s$, $\kappa_{\Gamma}$, and $\alpha$, with definitions given in the proof section.
\end{theorem}
See \ref{sec:proofOfTheorem_supportConsistency} for the proof.

\begin{remark}
	The order of estimation error bound result is equal to the result given by two datasets with $m$  and $m'$ samples drawn i.i.d. from two distinct non-paranormal models in the same senses.
\end{remark}

\begin{theorem}
\label{signConsistencyofDeltaEstimator}
Under the condition and notation  in Theorem \ref{errorBoundofDeltaEstimator}, if 
\begin{equation*}
    \min_{j,k : \Delta^*_{j,k} \neq 0} \left| \Delta^*_{j,k} \right| \ge 2 M_G \sqrt{\frac{\eta \log d + \log 2}{\min \left(m, m' \right)}}
\end{equation*}
for some $\eta>2$, then $M\left( \hat{\Delta} \right) = M\left( \Delta^* \right)$ with probability $1-2/d^{\eta-2}$.
\end{theorem}
See \ref{sec:proofOfTheorem_signConsistency} for the proof.

\section{Results}
\label{sec:Experiments}
In this section, we present experimental results showing the efficacy of our proposed approach in terms of speed and accuracy using synthetic data on one hand, and its applicability in inferring differential structures from real data on the other hand. 

\subsection{Synthetic data}

To generate synthetic data, we construct two subject-specific networks. The first network structure is chosen to be scale-free since it correctly models many real data \citep{barabasi2009scale}. The second network structure is constructed by randomly deleting and inserting 20 edges of the first network.
This process enables us to identify the zero entries in the precision matrices of each group.
Next, we generate non-zero entries of the precision matrices by sampling from $w = \textrm{sign}(v) + v$ where $v\sim \textrm{N}(0,1)$. If the matrices are not positive definite, $(1+\gamma) I$ with an appropriate $\gamma$ is added to the matrices to make them positive 
definite. 

Having constructed the covariance matrix $\Sigma$ for the first network, the data is generated by drawing i.i.d. samples from $\mathrm{NPN}\left(\mathbf{0},\Sigma,\mathbf{f}_{s}\right)$. To construct  $\mathbf{f}_{s}$, we randomly select $d$ functions from $\left\lbrace 2+x, 2x, 2^{x}, x^{3}, \sqrt[3]{x} \right\rbrace$. The data for the second network is obtained similarly by sampling from $\mathrm{NPN}\left(\mathbf{0},\Sigma',\mathbf{f}'_{s}\right)$ where $\Sigma'$ is the constructed covariance matrix and $\mathbf{f}'_s$ is generated similar to that of $\mathbf{f}_s$.

The accuracy is evaluated using the precision-recall curve. Let $\hat{\Delta}$ be an estimator of $\Delta^*$. The precision and the recall are defined as
\begin{align}
& \textrm{precision} = \frac{\sum_{i<j} \mathbbm{1} \lbrace\hat{\Delta}_{i,j} \neq 0 \And \Delta^*_{i,j} \neq 0 \rbrace}{\sum_{i<j} \mathbbm{1} \lbrace \hat{\Delta}_{i,j} \neq 0 \rbrace}, \nonumber \\
& \textrm{recall} = \frac{\sum_{i<j} \mathbbm{1} \lbrace\hat{\Delta}_{i,j} \neq 0 \And \Delta^*_{i,j} \neq 0 \rbrace}{\sum_{i<j} \mathbbm{1} \lbrace \Delta^*_{i,j} \neq 0 \rbrace}, \nonumber
\end{align}
where $\mathbbm{1} \lbrace . \rbrace$ is the indicator function.

In the first experiment, the performance of the solution path approach is compared with two existing approaches, namely the D-trace approach \citep{yuan2017differential} and CrossFDTL approach \citep{wu2020differential}. 
Dtrace function is used from the R package DiffGraph \citep{zhang2018diffgraph} for D-trace method implementation.
We refer to this method as APGD D-trace because it employs the accelerated proximal gradient descent technique.
Our proposed method and CrossFDTL method are implemented using the R package Rcpp \citep{eddelbuettel2011rcpp} in C++.

\begin{figure*} [t!]
    \centering
    \includegraphics[width=1\linewidth]{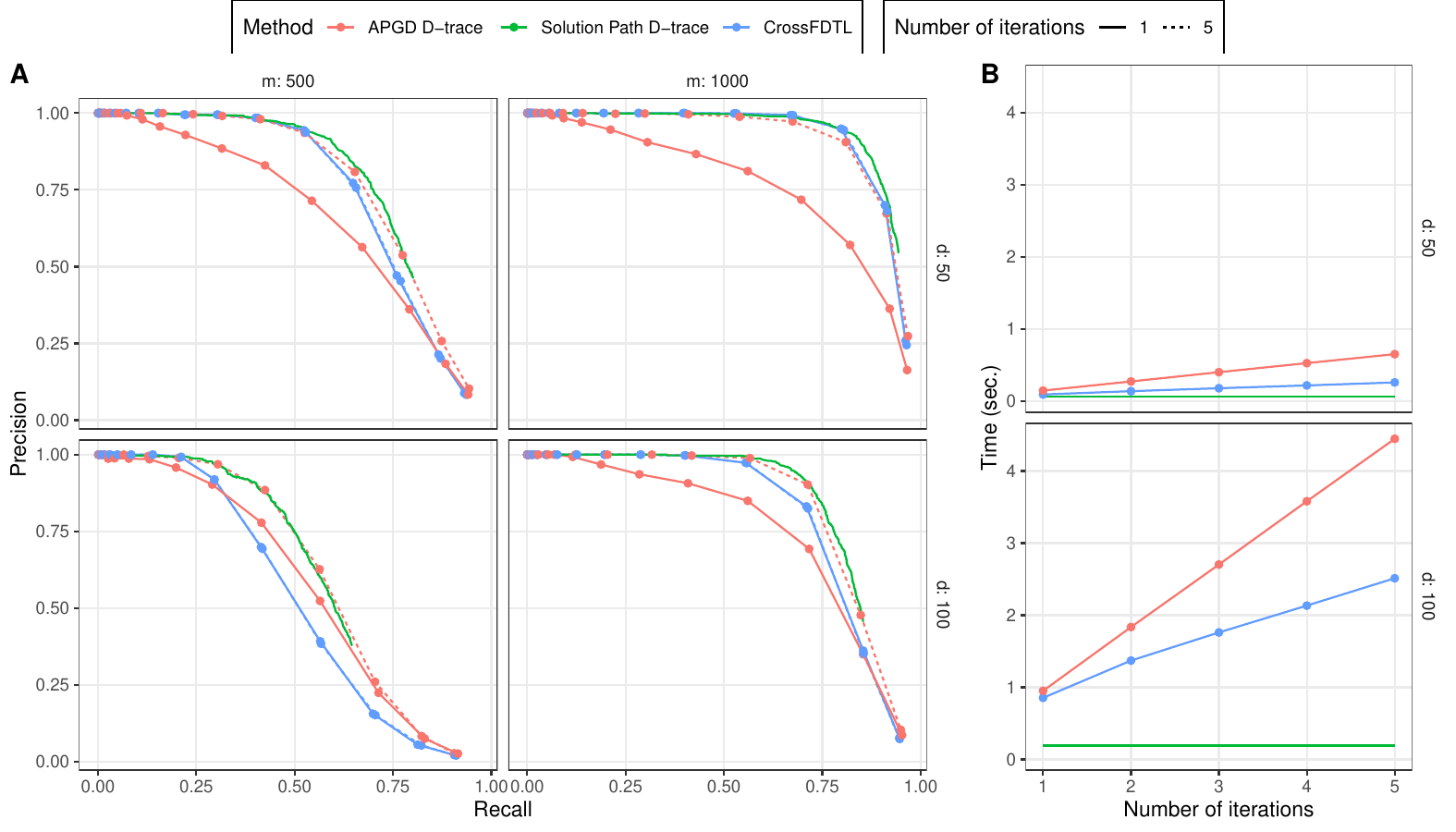}
    \caption{Performance of different approaches on synthetic data with 20 differential edges between two subjects, and different combinations of the number of iterations. The lines color and type indicate the method type and the number of iterations in iterative methods, respectively. Also, each point corresponds to the solution of an iterative method for a specific value of $\lambda$.
    (\textbf{A}) Accuracy of different method.
    (\textbf{B}) Speed of different methods.
    }
    \label{fig:SPD-trace vs SVD-trace}
\end{figure*}

Figure \ref{fig:SPD-trace vs SVD-trace} depicts the comparison of the precision-recall curves and computational time of the three methods averaged over 100 random generations of the data, with $d \in \{50, 100\}$ and $m = m' \in \{ 500, 1000 \}$. In each plot, 50 different values from $\left[ 0.1, 2\right]$ are chosen for the APGD D-trace
and CrossFDTL methods. To make a fair comparison between the methods, $c=100$ is set to control the range of the regularization parameter values. We experimentally observed that our proposed method identifies around 50 knots in the solution path (similar to the number of different tuning parameter values).
Moreover, the CrossFDTL method has an additional tuning parameter ($\rho$) that controls the Frobenius norm of $\hat{\Delta}$. For a fair comparison, we set $\rho = 0$ in experiments.

We observe that the precision-recall curves of our proposed method outperforms the accuracy curves of APGD D-trace and CrossFDTL methods.
The difference between the accuracy curves of our proposed method and APGD D-trace method
is due to the fact that our proposed method detects all sparse differential network that can be obtained for a specific range of regularization parameter values, while the APGD D-trace method ignores many high-precision sets due to its shortcoming in finding the solution path. In other words, if we generate a curve by performing APGD D-trace with a sufficient number of iterations for all required values of $\lambda$, the obtained curve will be similar to the Solution Path D-trace curve.
Obviously, the two methods coincide at $\lambda$s chosen for the APGD D-trace with 5 iterations. In addition, we observe that our proposed method is much faster than the APGD D-trace and CrossFDTL methods. More concretely, our proposed method solves the problem for the more number of tuning parameter values in a more accurate manner with a less consumption of time.

\begin{figure} [t!]
    \centering
    \includegraphics[width=0.7\linewidth]{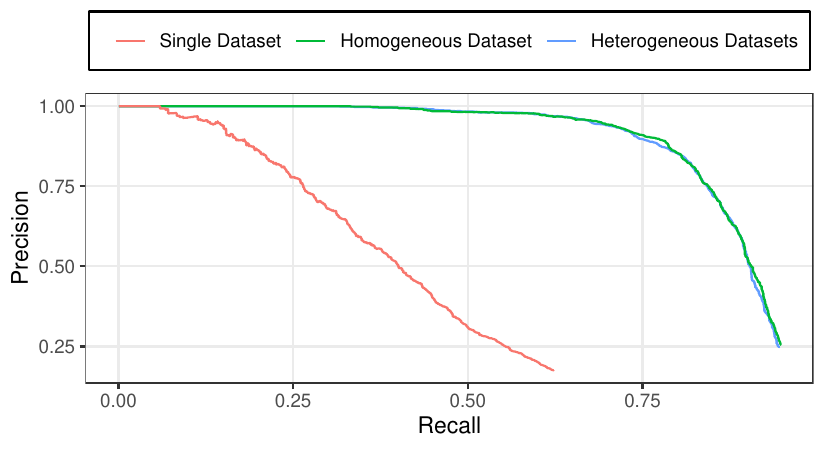}
    \caption{Precision-recall curves of the single dataset, homogeneous dataset, and heterogeneous datasets.}
    \label{fig:Integration Effects}
\end{figure}

Our approach leverages the combination of heterogeneous datasets to obtain a reliable estimation of the covariance matrix. To demonstrate the effectiveness of this approach, 
we conduct the second experiment in which we utilize four different datasets per group, with each dataset containing 200 samples. In Figure \ref{fig:Integration Effects}, the precision-recall curve averaged over 100 random generations is drawn (the blue curve). If only one dataset is used for the inference, then the performance is dramatically degraded as is shown in the figure (the red curve). Finaly, if the dataset were assumed to be homogeneous then we had access to a single dataset with size 800 samples. The green curve depicting the performance of the homogeneous case. We observe that the efficacy curve of the heterogeneous scenario, despite the presence of different datasets, is similar to that of the homogeneous scenario with the same sample size. This similarity suggests that our strategy efficiently exploits the information available in the heterogeneous datasets.

\subsection{Real data}
\label{sec:Real data application}

We consider gene expression datasets of ovarian cases in TCGA \citep{cancer2011integrated} with two groups of cases classified based on drug responses using the criterion proposed by \cite{nabavi2016emdomics}. We have selected 242 platinum-sensitive tumors and 98 platinum-resistant ones. In order to improve the learning time and model interpretability, the 551 genes of relevant pathways proposed by \cite{dilruba2016platinum, burris2013overcoming} are considered.

Our proposed method results in a solution path, i.e., a collection of different solutions. Hence, the selection of a solution from the resulting collection is critical. To this end, we generate a random collection of datasets with 80 percent of samples in both subjects. By applying our proposed method to the obtained random datasets, a solution path is extracted. We repeat this procedure 10 times and compute the instability of the results for different values of the regularization parameter. By setting the instability threshold equal to 0.001, we identify the appropriate regularization parameter value and corresponding differential network inferred by all data. It is worth mentioning that our strategy is adopted from the StARS method \citep{liu2010stability}.

\begin{figure} [t!]
    \centering
    \includegraphics[width=1\linewidth]{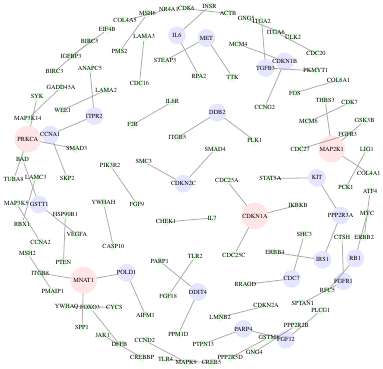}
    \caption{The inferred differential network between platinum-sensitive and platinum-resistant ovarian tumors with the regularization parameter are obtained by StARS method. The node size and color correspond to its degree.}
    \label{fig:structural changes of platinum resistant}
\end{figure}

 Figure \ref{fig:structural changes of platinum resistant} depicts the obtained differential network. It contains 77 edges among 127 genes. Notably, we found supportive evidence for all genes that have the most degree in the identified network including CDKN1A \citep{xia2011cytoplasmic, lincet2000p21cip1,koster2010cytoplasmic}, MAP2K1 \citep{penzvalto2014mek1}, MNAT1 \citep{zhou2018mnat1,qiu2020mnat1}, and PRKCA \citep{mohanty2005enhancement, arrighetti2016pkc,basu1995comparison}. Furthermore, we obtained 912 genes associated with platinum resistance in cancer from the database introduced by \cite{huang2021highly}. We consider two groups of genes in the identified differential network including 1) genes involved in at least one edge (non-isolated genes), and 2) genes with a degree greater than 1 (connector genes). According to Fisher's exact test, non-isolated and connector genes are significantly enriched with platinum resistance-related genes (p-values are 0.01 and 0.003, respectively).

\section{Conclusions}
\label{sec:Conclusion}
In this paper, learning of sparse differential network between two classes of non-paranormal graphical models is addressed. We assume a number of heterogeneous datasets are given as input, where datasets of each class are drawn from different NPN models, but with the same underlying covariance matrix. We consider the minimization of the lasso penalized D-trace loss function \citep{yuan2017differential} to identify differential network. In this regard, it has been proved that the solution path of the D-trace model \citep{yuan2017differential} is continuous piecewise linear with respect to the regularization parameter. We propose an efficient method for extracting the exact solution path and integrating heterogeneous datasets in covariance matrix estimation. Our proposed method improves the speed and accuracy of detecting differential network. The competitive advantages of our proposed approach are demonstrated using synthetic data. In addition, we evaluate the efficacy of our method in a real application as well. By applying our method to gene expression data of ovarian cancer, we identified drug resistance-related genes which are already validated by independent studies. Also, our approach can be extended in several directions. For instance, it can be applied to the model proposed by \cite{zhang2014sparse} for sparse precision matrix estimation in NPN distributions. As another extension, one can study and solve a similar problem with ordered weighted $\ell_1$ regularization.

\begin{appendix}
\label{section_Proof}

\section{Proof of Theorem \ref{piecewise linear solution path}}
\label{sec:Proof of Theorem piecewise linear solution path}
Inspired by \cite{rosset2007piecewise}, we establish  the theorem using the subgradient optimality conditions of the optimization problem \eqref{lasso penalized D-trace loss function minimization}.  It is worth noting that the lasso penalized D-trace loss function is a convex function if the estimated covariance matrices are positive semi-definite \citep{yuan2017differential}. Consequently, we can use the subgradient optimality to find the optimal solution. Hence, we have
	
	\begin{align}
	0_{d,d} &\in \partial \left(
	L_D\left(\hat{\Delta}\left(\lambda\right);
	\hat{\Sigma},
	\hat{\Sigma}'\right) +
	\lambda \norm{\hat{\Delta}\left(\lambda\right)}_{1}
	\right) \nonumber \\
	&= 
	\frac{1}{2}
	\left(
	\hat{\Sigma}
	\hat{\Delta}\left(\lambda\right)
	\hat{\Sigma}' + 
	\hat{\Sigma}'
	\hat{\Delta}\left(\lambda\right)
	\hat{\Sigma}
	\right)- \hat{\Sigma} +
	\hat{\Sigma}' +
	\lambda \partial \norm{\hat{\Delta}\left(\lambda\right)}_{1}, \nonumber
	\end{align}
	where $\hat{\Delta}\left(\lambda\right)$ is the minimizer of the optimization problem \eqref{lasso penalized D-trace loss function minimization}.
	We rewrite the above relation in the vectorized form as
	    $\vec{0} \in 
	\left\lbrace
	\Gamma \vec{\Delta}\left(\lambda\right)
	+ \mathbf{v} + \lambda \partial \norm{\vec{\Delta}\left(\lambda\right)}_{1}
	\right\rbrace$,
	where $\Gamma = \frac{1}{2} \left \lbrace \hat{\Sigma} \otimes \hat{\Sigma}' +
	\hat{\Sigma}' \otimes \hat{\Sigma} \right \rbrace$,
	$\mathbf{v} = \textrm{vec}\left(
	\hat{\Sigma}' - \hat{\Sigma}
	\right)$,
	$\vec{\mathbf{0}} = \textrm{vec}\left(0_{d,d}\right)$ and $\vec{\Delta}\left(\lambda\right) = \textrm{vec}\left(\hat{\Delta}\left(\lambda\right)\right)$.
	Therefore, for some $\mathbf{\nu} \in \partial \norm{\vec{\Delta}\left(\lambda\right)}_{1}$, i.e.,
	\begin{equation*}
	    \nu_{i} \in
	\begin{cases}
	\left\lbrace 1 \right\rbrace, &\text{if $\vec{\Delta}\left(\lambda\right)_{i} > 0$}\\
	\left\lbrace -1 \right\rbrace, &\text{if $\vec{\Delta}\left(\lambda\right)_{i} < 0$}\\
	\left[-1,1\right], &\text{if $\vec{\Delta}\left(\lambda\right)_{i} = 0$}
	\end{cases}, i = 1,\dots,d^2,
	\end{equation*}
	we have
	$\Gamma \vec{\Delta}\left(\lambda\right)
	+ \mathbf{v} = -\lambda \mathbf{\nu}$. Thus, we can derive the optimality conditions for any $i \in \left\lbrace 1, \dots, d^2\right\rbrace$ as
	\begin{align}
	\label{subgradient optimality conditions}
	\begin{cases}
	\Gamma_{i,:} \vec{\Delta}\left(\lambda\right)
	+ \mathbf{v}_i = -\lambda \textrm{sign}  \left(\vec{\Delta}\left(\lambda\right)_{i}\right), &\text{if $\vec{\Delta}\left(\lambda\right)_{i} \ne 0$}\\
	\lvert \Gamma_{i,:} \vec{\Delta}\left(\lambda\right)
	+ \mathbf{v}_i \rvert \le \lambda, &\text{if $\vec{\Delta}\left(\lambda\right)_{i} = 0$.}
	\end{cases}
	\end{align}
	
	For every value of $\lambda$, we have a set of "active" variables $\mathcal{A}_\lambda$ that have nonzero coefficients, i.e., $\mathcal{A}_\lambda=\left\lbrace i:\vec{\Delta}\left(\lambda\right)_{i} \ne 0 \right\rbrace$. If we have an active set $\mathcal{A}_{\lambda}$ and the corresponding active signs $\mathbf{s}_{\mathcal{A}_{\lambda}} = \textrm{sign}\left( \vec{\Delta}\left(\lambda\right)_{\mathcal{A}_{\lambda}} \right)$,
 the problem becomes a system of linear equation. If the matrix $\Gamma_{\mathcal{A}_{\lambda}, \mathcal{A}_{\lambda}}$ is invertible, then the system has a unique solution given by
	\begin{align}
	\label{delta computation formula}
	&\vec{\Delta}\left(\lambda\right)_{\mathcal{A}_{\lambda}} = -\left(\Gamma_{\mathcal{A}_{\lambda}, \mathcal{A}_{\lambda}}\right)^{-1} \left(\mathbf{v}_{\mathcal{A}_{\lambda}} + \lambda \mathbf{s}_{\mathcal{A}_{\lambda}}\right), \\
&\vec{\Delta}\left(\lambda\right)_{\mathcal{A}_{\lambda}^{c}} = 0.
	\end{align}
 
	As we decrease the value of $\lambda$, the optimality conditions remain satisfied as long as the corresponding active set is unchanged. Hence, there is an interval of tuning parameter values that equation \eqref{delta computation formula} is maintained for all $\lambda \in \left( \lambda_{t+1},\lambda_t\right)$, where $\lambda_t$ and $\lambda_{t+1}$ are the limit points of this interval. For sake of convenience, we denote the corresponding active set for all the regularization parameter values $\lambda \in \left(\lambda_{t+1},\lambda_t\right)$ as $\mathcal{A}_{\lambda_t}$. Thus, for all $\lambda \in \left(\lambda_{t+1},\lambda_t\right)$, we have
	\begin{align}
	\label{delta computation formula using knot}
	&\vec{\Delta}\left(\lambda\right)_{\mathcal{A}_{\lambda_t}} = -\left(\Gamma_{\mathcal{A}_{\lambda_t}, \mathcal{A}_{\lambda_t}}\right)^{-1} \left(\mathbf{v}_{\mathcal{A}_{\lambda_t}} + \lambda \mathbf{s}_{\mathcal{A}_{\lambda_t}}\right), \\
&\vec{\Delta}\left(\lambda\right)_{\mathcal{A}_{\lambda_t}^{c}} = 0.
	\end{align}
Therefore, there are $\infty = \lambda_0 > \lambda_1 > \lambda_2 > \dots > \lambda_T$, such that the matrix $\Gamma_{\mathcal{A}_{\lambda_t}, \mathcal{A}_{\lambda_t}}$ is invertible and $\hat{\Delta}\left(\lambda\right)$ for any $\lambda \in \left(\lambda_{t+1},\lambda_t\right)$ is given by \eqref{delta computation formula using knot}. Thus, we require to identify the values of the tuning parameter that the corresponding active set is changed, along with corresponding solutions, i.e., $\hat{\Delta}\left(\lambda_{t}\right)$ for all $t \in \left\lbrace 0, 1, \dots, T\right\rbrace$. We call these points knot, and these changes are generated when some variables enter to active set or leave it.
	
We first consider the case that some variables leave the active set. When some variables leave the active set, their corresponding values become zero. In other words, suppose that $\lambda_{t}$ is determined and $\lambda^{cross}$ represents the value of the next knot generated due to some variables leaving the active set. The optimality conditions \eqref{delta computation formula using knot} imply that $v_{t} \in \mathcal{A}_{\lambda_{t}}$ leaves the active set at $\lambda^{cross}$ if
	\begin{align}
	\label{crossing time condition}
	- \left[
	\left(\Gamma_{\mathcal{A}_{\lambda_{t}}, \mathcal{A}_{\lambda_{t}}}\right)^{-1}
	\right]_{v_{t},:} \left(\mathbf{v}_{\mathcal{A}_{\lambda_{t}}} + \lambda^{cross} \mathbf{s}_{\mathcal{A}_{\lambda_{t}}}\right) =0.
	\end{align}
	Solving \eqref{crossing time condition} yields
	\begin{equation*}
	    \lambda^{cross} = 
	\frac{- \left[
		\left(\Gamma_{\mathcal{A}_{\lambda_{t}}, \mathcal{A}_{\lambda_{t}}}\right)^{-1}
		\right]_{v_{t},:} \mathbf{v}_{\mathcal{A}_{\lambda_{t}}}}{ \left[
		\left(\Gamma_{\mathcal{A}_{\lambda_{t}}, \mathcal{A}_{\lambda_{t}}}\right)^{-1}
		\right]_{v_{t},:} \mathbf{s}_{\mathcal{A}_{\lambda_{t}}}}.
	\end{equation*}
	It is obvious that the value of the next knot is the nearest possible value to the current knot value. Therefore, we have
	\begin{align}
	v_t= \arg \max_{v}
	\quad 
	&	C\left(v\right), \nonumber \\
	\textrm{s.t.} \quad & C\left(v\right) < \lambda_{t}, v \in \mathcal{A}_{\lambda_{t}} \nonumber
	\end{align}
	where $C\left(v\right) =
	    \frac{- \left[
		\left(\Gamma_{\mathcal{A}_{\lambda_{t}}, \mathcal{A}_{\lambda_{t}}}\right)^{-1}
		\right]_{v,:} \mathbf{v}_{\mathcal{A}_{\lambda_{t}}}}{ \left[
		\left(\Gamma_{\mathcal{A}_{\lambda_{t}}, \mathcal{A}_{\lambda_{t}}}\right)^{-1}
		\right]_{v,:} \mathbf{s}_{\mathcal{A}_{\lambda_{t}}}}$.
  
	Consequently, there are no leaving events for any value of $\lambda \in \left(C\left(v_t\right), \lambda_{t}\right)$.
	If the leaving variable is not excluded from the active set at $C\left(v_t\right)$ and the new knot value is not calculated, then the optimality conditions \eqref{subgradient optimality conditions} are violated for any $\lambda < C\left(v_t\right)$ due to the violation of \eqref{delta computation formula using knot}.
	
	Now we consider the case that some variables enter the active set. Suppose that $\lambda^{hit}$ is the value of the next knot generated by entering some variables. The optimality conditions \eqref{subgradient optimality conditions} and  \eqref{delta computation formula using knot} imply that there is a $v'_{t} \in \mathcal{A}^{c}_{\lambda_{t}}$ such that
	\begin{align}
	\label{hit time conditions}
	\Gamma_{v'_{t},\mathcal{A}_{\lambda_{t}}} 
	\left(\Gamma_{\mathcal{A}_{\lambda_{t}}, \mathcal{A}_{\lambda_{t}}}\right)^{-1} \left(\mathbf{v}_{\mathcal{A}_{\lambda_{t}}} + \lambda_{t} \mathbf{s}_{\mathcal{A}_{\lambda_{t}}}\right)
	- \mathbf{v}_{v'_{t}} = s_{v'_{t}}\lambda_{t}^{hit}.
	\end{align}
	By solving \eqref{hit time conditions}, we obtain
	\begin{equation*}
	    \lambda^{hit} = \frac{\Gamma_{v'_{t},\mathcal{A}_{\lambda_{t}}} \left(\Gamma_{\mathcal{A}_{\lambda_{t}}, \mathcal{A}_{\lambda_{t}}}\right)^{-1} \mathbf{v}_{\mathcal{A}_{\lambda_{t}}}- \mathbf{v}_{v'_{t}}}
	{s_{v'_{t}} - \Gamma_{v'_{t},\mathcal{A}_{\lambda_{t}}} \left(\Gamma_{\mathcal{A}_{\lambda_{t}}, \mathcal{A}_{\lambda_{t}}}\right)^{-1} \mathbf{s}_{\mathcal{A}_{\lambda_{t}}}}.
	\end{equation*}
	It is obvious that the value of the next knot is the nearest possible value to the current knot value. Hence, we have
	\begin{align}
	v'_t= \arg \max_{v'}
	\quad 
	&	H\left(v'\right), \nonumber \\
	\textrm{s.t.} \quad & H\left(v'\right) < \lambda_{t}, v' \in \mathcal{A}^{c}_{\lambda_{t}}, s_{v'} \in \left \lbrace -1, 1 \right \rbrace \nonumber
	\end{align}
	where $H\left(v'\right)  =
	\frac{\Gamma_{v',\mathcal{A}_{\lambda}} \left(\Gamma_{\mathcal{A}_\lambda, \mathcal{A}_\lambda}\right)^{-1} \mathbf{v}_{\mathcal{A}_\lambda}- \mathbf{v}_{v'}}
	{s_{v'} - \Gamma_{v',\mathcal{A}_{\lambda}} \left(\Gamma_{\mathcal{A}_\lambda, \mathcal{A}_\lambda}\right)^{-1} \mathbf{s}_{\mathcal{A}_\lambda}}$.
	
	Therefore, no entering event occurs for any $\lambda \in \left(H\left(v'_t\right), \lambda_{t}\right)$.
	If the entering variables are not involved in the active set at $H\left(v'_t\right)$, then the optimality conditions \eqref{subgradient optimality conditions} are violated for any $\lambda < H\left(v'_t\right)$, as \eqref{delta computation formula using knot} is not satisfied.
	
 By considering two possible cases, the tuning parameter value of the next knot must be equal to the nearest possible event to maintain optimality conditions \eqref{subgradient optimality conditions} using \eqref{delta computation formula using knot}, i.e., $\lambda_{t+1} = \max \left\lbrace C\left(v_t\right), H\left(v'_t\right) \right\rbrace$. Also, the final answer obtained by placing $\mathcal{A}_{\lambda_{t}}$ in \eqref{delta computation formula using knot} is equal to placing $\mathcal{A}_{\lambda_{t+1}}$ in \eqref{delta computation formula using knot} because the value of $\lambda_{t+1}$ implies that the value of variables entered or exited from the active set is equal to zero. Thus, we can compute $\hat{\Delta} \left(\lambda\right)$ for all $\lambda \in \left[ \lambda_{t+1}, \lambda_{t} \right]$ using \eqref{delta computation formula using knot} if we know $\mathcal{A}_{\lambda_{t}}$ and $\mathbf{s}_{\mathcal{A}_{\lambda_t}}$. Hence, we can compute $\hat{\Delta} \left( \lambda \right)$ only using $\hat{\Delta} \left(\lambda_{t+1}\right)$ and $\hat{\Delta}\left( \lambda_{t} \right)$ extracted from \eqref{delta computation formula using knot} with some algebra. In other words, for all $\lambda \in \left[ \lambda_{t+1}, \lambda_{t} \right]$ we have
	\begin{align}
	\label{delta computation formula using knot (simplified version)}
	&\hat{\Delta}\left(\lambda\right)_{\mathcal{A}_{\lambda_t}} =
	\frac{\lambda_{t} - \lambda}{\lambda_{t} - \lambda_{t+1}} \hat{\Delta} \left( \lambda_{t+1} \right)_{\mathcal{A}_{\lambda_t}} +
	\frac{\lambda - \lambda_{t+1}}{\lambda_{t} - \lambda_{t+1}} \hat{\Delta} \left( \lambda_{t} \right)_{\mathcal{A}_{\lambda_t}}, \nonumber \\
	&\hat{\Delta}\left(\lambda\right)_{\mathcal{A}_{\lambda_t}^{c}} = 0.
	\end{align}
	
	Let $\lambda \ge 0$ be an arbitrary value. Assume that there are knot points, i.e., $\infty = \lambda_0 > \lambda_1 > \lambda_2 > \dots > \lambda_T$, such that $\hat{\Delta}\left(\lambda\right) = 
	\frac{\lambda_{t} - \lambda}{\lambda_{t} - \lambda_{t+1}} \hat{\Delta} \left( \lambda_{t+1} \right) +
	\frac{\lambda - \lambda_{t+1}}{\lambda_{t} - \lambda_{t+1}} \hat{\Delta} \left( \lambda_{t} \right)$ for $\lambda \in \left[\lambda_{t+1},\lambda_{t}\right]$. If $\lambda \in [\lambda_{1}, \lambda_0)$ then it is obvious that $\hat{\Delta} \left( \lambda \right) = \mathbf{0}_{d,d}$. Also, we can compute $\mathcal{A}_{\lambda_{t+1}}$, $\mathbf{s}_{\mathcal{A}_{\lambda_{t+1}}}$, and $\vec{\Delta}\left(\lambda\right)$ for all $\lambda \in \left[\lambda_{t+1}, \lambda_{t}\right]$ using \eqref{delta computation formula using knot (simplified version)} if we have active set $\mathcal{A}_{\lambda_t}$ and active sign $\mathbf{s}_{\mathcal{A}_{\lambda_t}}$. Therefore, there are $\infty = \lambda_0 > \lambda_1 > \lambda_2 > \dots > \lambda_T$, such that	$\hat{\Delta}\left(\lambda\right)$ for any $\lambda \in \left(\lambda_T,\infty\right)$ is given by  
	\begin{equation*}
	    \frac{\lambda_{t} - \lambda}{\lambda_{t} - \lambda_{t+1}} \hat{\Delta} \left( \lambda_{t+1} \right) +
	\frac{\lambda - \lambda_{t+1}}{\lambda_{t} - \lambda_{t+1}} \hat{\Delta} \left( \lambda_{t} \right), \forall \lambda \in \left[\lambda_{t+1}, \lambda_t\right].
	\end{equation*}

\section{Proof of Lemma \ref{errorBoundOfCorrelationEstimator}}
\label{lem:app_proof}

    The lemma is proved by applying McDiarmid’s inequality. We first show the bounded difference property for the function $\phi: \left(\mathbb{R} \times \mathbb{R}\right)^m \mapsto \mathbb{R}$ defined as 
        $\phi \left(
			\left\lbrace
				X_{1,k}^{\left(1\right)}, X_{1,l}^{\left(1\right)}
			\right\rbrace,
			\dots,
			\left\lbrace
			X_{m_N,k}^{\left(N\right)}, X_{m_N,l}^{\left(N\right)}
			\right\rbrace
		\right) = 
		\left[\hat{\tau}_{w}\right]_{k,l}$.
Let
 $z_{s,i} = \left\lbrace
	X_{i,k}^{\left(s\right)}, X_{i,l}^{\left(s\right)}
	\right\rbrace$
 and 
 $h \left( z_{s,i}, z_{s,j} \right) = \textrm{sign}\left(
	\left( X_{i,k}^{\left(s\right)}-X_{j,k}^{\left(s\right)} \right)
	\left( X_{i,l}^{\left(s\right)}-X_{j,l}^{\left(s\right)} \right) \right)$.
 For any $s \in \left[N\right], i \in \left[m_{s}\right]$ and any $z_{1,1}, \dots, z_{N,m_N}, z_{s,i}' \in \left( \mathbb{R} \times \mathbb{R} \right)$, we have 
	\begin{align}
	&|	\phi \left(\dots,z_{s,i},\dots,\right)
	- \phi \left(\dots,z_{s,i}',\dots,\right)| \nonumber = \left| \left[\hat{\tau}_{w}\right]_{k,l} - \left[\hat{\tau}_{w}\right]_{k,l}' \right|\nonumber \\
	&= \left| \sum_{s=1}^{N} \alpha_s \left[\hat{\tau}_{s}\right]_{k,l} -
	\sum_{s=1}^{N} \alpha_s \left[\hat{\tau}_{s}\right]_{k,l}' \right| \nonumber \\
	&= \frac{2\alpha_s}{m_s \left(m_s -1\right)} | \sum_{j \ne i} h \left( z_{s,i}, z_{s,j} \right) - h \left( z_{s,i}', z_{s,j} \right) | \nonumber \\
	&\le \frac{2\alpha_s}{m_s \left(m_s -1\right)} \times 2\left(m_s -1\right)
	= \frac{4 \alpha_s}{m_s}. \nonumber
	\end{align}
	
McDiarmid's Inequality gives
	\begin{align}
	&\mathbb{P}\left( \left[\hat{\tau}_{w}\right]_{k,l} - \left[\tau_{w}\right]_{k,l} \ge t\right)
	 \le
	\exp\left(\frac{-2t^2}{\sum_{s=1}^{N}\sum_{i=1}^{m_s}\left(\frac{4 \alpha_s}{m_s}\right)^2}\right) = \exp \left(\frac{-t^2}{8 \sum_{s=1}^{N} \frac{{\alpha_s}^2}{m_s}}\right). \nonumber 
	\end{align}
	We set $\alpha_s = \frac{m_s}{m}$ for all $s \in \left[N\right]$ to obtain the tightest bound. Hence, we have
	\begin{equation}
	    \mathbb{P}\left( \left[\hat{\tau}_{w}\right]_{k,l} - \left[\tau_{w}\right]_{k,l} \ge t\right) \le 
	\exp \left(\frac{-mt^2}{8}\right).
	\label{eq:kendall's tau PAC}
	\end{equation}
	
	The upper bound of
	$\mathbb{P}\left(
	\left[\tau_{w}\right]_{k,l} -
	\left[\hat{\tau}_{w}\right]_{k,l} \ge t\right)$ is obtained by similar arguments. By applying the union bound, we have
	
	\begin{equation*}
		\mathbb{P}\left(\left| \left[\hat{\tau}_{w}\right]_{k,l} - \left[\tau_{w}\right]_{k,l} \right|\ge t\right) \le
		2\exp \left(\frac{-mt^2}{8}\right).
	\end{equation*}
	
	To obtain the error bound for our proposed correlation estimator, we use the Lipschitz continuity property of the $\textrm{sin}(.)$ function. Hence, we have
	\begin{align*}
	&
	\mathbb{P}\left(
	\left|\tilde{\Sigma}_{k,l} - {\Sigma}_{k,l}\right| \ge t
	\right) 
    \stackrel{(a)}{=}
	\mathbb{P}\left( 
	\left|
	\sin\left(\frac{\pi}{2}
	\left[\hat{\tau}_{w}\right]_{k,l}
	\right) -
	\sin\left(\frac{\pi}{2}
	\left[\tau\right]_{k,l}
	\right)
	\right| \ge \frac{t}{4}
	\right) \\
	&\stackrel{(b)}{\le}
	\mathbb{P}\left(
	\left|
	\frac{\pi}{2}
	\left[\hat{\tau}_{w}\right]_{k,l} -
	\frac{\pi}{2}
	\left[\tau_{w}\right]_{k,l}
	\right| \ge t
	\right) \stackrel{(c)}{\le}
	2 \exp \left(
	\frac{- m t^2}{2\pi^2}
	\right),
	\end{align*}
where $(a)$ comes from \eqref{Greiner's equality}, $(b)$ comes from the fact that $\textrm{sin}\left(.\right)$ is a $1$-Lipschitz function, i.e.,
	$\left|
	\sin \left(x\right) - \sin \left(y\right)
	\right| \le \left|x-y\right|$,
	and $(c)$ is obtained from the derived inequality in  \eqref{eq:kendall's tau PAC}.

\section{Proof of Theorem \ref{errorBoundofDeltaEstimator}}
\label{sec:proofOfTheorem_supportConsistency}

We follow the methodology used by \cite{yuan2017differential} for the proof of Theorem 1. In this methodology, the conclusion of Theorem \ref{errorBoundofDeltaEstimator} is obtained based on the performance of the covariance estimation. Data model and covariance estimation method are the main difference between Theorem 1 presented by \cite{yuan2017differential} and our proposed Theorem \ref{errorBoundOfCorrelationEstimator}. Hence, it is required to prove that the performance of our proposed covariance estimator is similar to the performance of the covariance estimator proposed by \cite{yuan2017differential}. More concretely, it is required to prove that there exists a constant $\nu_* > 0$ and a function $f: \mathbb{N} \times (0,\infty) \mapsto (0,\infty)$ such that
\begin{equation*}
    \mathbb{P}\left(
	\left|\tilde{\Sigma}_{k,l} - {\Sigma}_{k,l}\right| \ge t
	\right) \le 1 / f(m,t)
\end{equation*}
for all $k,l \in [d]$ and $0<t<1/\nu_*$. Based on Lemma \ref{errorBoundOfCorrelationEstimator}, the condition is satisfied with $\nu_* = 0$ and $f(m,t) = 0.5 \exp{\left( \frac{mt^2}{2\pi^2} \right)}$ in our proposed method. Also, we can set $M = 1$ since we assumed that the covariance matrices $\Sigma$ and $\Sigma'$ are correlation matrices, i.e, $\max \left[ \norm{\Sigma}_{\infty}, \norm{\Sigma'}_{\infty} \right] \leq 1$. Now, we substitute $\nu_*$ and $f(m,t)$ in the proof of Theorem 1 presented by \cite{yuan2017differential}. The substitution changes some parameter definitions and we have

\begin{align*}
&\bar{\delta} = \min \left\lbrace
-1 + \sqrt{1+ \left(6s \kappa_{\Gamma} \right)^{-1}},
-1 + \sqrt{2 + \tilde{M}^{-1}},
A
\right\rbrace, \\
&M_G = \sqrt{2} \pi \kappa_{\Gamma}  \left( C_{G1} + C_{G2} \right),
n_f \left(
\bar{\delta}, d^{\eta}
\right) = 2{\pi}^2 \bar{\delta}^{-2} \left( \eta \log d + \log 2 \right),
\end{align*}
where
\begin{align*}
& A = \frac{\alpha}{4-\alpha}, C_{G2} = 6  s \kappa_{\Gamma} \left( 2 + A \right), \\
&C_{G1} =  \left( 1 + 3s \kappa_{\Gamma}A \left( A + 2 \right) \right)
\left(
2  + \max \left[
\tilde{M} \left( 2 + A \right),
4  A^{-1}
\right]
\right).
\end{align*}

In addition, 
we employ the method proposed by \cite{zhao2014positive} to project $\tilde{\Sigma}$ onto the positive semi-definite cone, resulting in the estimation of $\hat{\Sigma}$.
From equation (A.6) presented by \cite{zhao2014positive}, we have
\begin{align}
\label{eq:projectionEffect}
        \norm{\hat{\Sigma} - \Sigma}_{\infty} \le
        2\norm{\tilde{\Sigma} - \Sigma}_{\infty} + \frac{\mu}{2},
\end{align}
where $\mu$ is an arbitrary parameter that controls the approximation error. On the other hand, the proof of Theorem 1 presented by \cite{yuan2017differential} is based on a concentration bound for the covariance matrix estimator. Specifically, for $m > n_f \left( \bar{\delta}, d^{\eta} \right)$, we have
\begin{align*}
    \mathbb{P}\left(
    \norm{ \tilde{\Sigma} - {\Sigma} }_{\infty} < \bar{\delta}
	\right) \ge 1-\frac{1}{d^{\eta}}.
\end{align*}
By setting $\mu = \bar{\delta}$ and incorporating \eqref{eq:projectionEffect}, for $m > n_f \left( \bar{\delta}, d^{\eta} \right)$, we have
\begin{align*}
    \mathbb{P}\left(
    \norm{ \hat{\Sigma} - {\Sigma} }_{\infty} < \frac{5}{2}\bar{\delta}
	\right) \ge 1-\frac{1}{d^{\eta}}.
\end{align*}
If we set $\bar{\bar{\delta}} = \frac{5}{2}\bar{\delta}$, we have $n_f \left( \bar{ \bar{\delta} }, d^{\eta} \right) = \frac{25}{2}{\pi}^2 \bar{\bar{\delta}}^{-2} \left( \eta \log d + \log 2 \right)$. Therefore, we have the basic concentration bound for projected estimator $\hat{\Sigma}$ and we can complete the proof of Theorem \ref{errorBoundofDeltaEstimator}. More concretely, for $m>\frac{25}{4}{\gamma}^2 \bar{\bar{\delta}}^{-2}$, where $\gamma= \pi\sqrt{2 \left(\eta \log{d} + \log{2} \right)}$, we have
\begin{align*}
    \mathbb{P}\left(
    \norm{ \hat{\Sigma} - {\Sigma} }_{\infty} < \bar{\bar{\delta}}
	\right) \ge 1-\frac{1}{d^{\eta}}.
\end{align*}

\section{Proof of Theorem \ref{signConsistencyofDeltaEstimator}}
\label{sec:proofOfTheorem_signConsistency}

The proof is similar to the proof of Theorem 2 presented by \cite{yuan2017differential}. We know from Theorem \ref{errorBoundofDeltaEstimator} that the nonzero elements of $\hat{\Delta}$ are a subset of the nonzero elements of $\Delta^*$. Moreover, we derive a probabilistic upper bound on the estimation error. In addition, it is assumed that the minimum absolute value of the nonzero elements of $\Delta^*$ is greater than twice the obtained probabilistic upper bound in Theorem \ref{errorBoundofDeltaEstimator}. Consequently, 
we can deduce the conclusion of Theorem \ref{errorBoundofDeltaEstimator}.
\end{appendix}


\bibliographystyle{imsart-number} 
\bibliography{paper}       

\end{document}